\documentclass[letterpaper, 10 pt, conference]{ieeeconf}
\IEEEoverridecommandlockouts
\usepackage{cite}
\usepackage{amsmath,amssymb,amsfonts}
\usepackage{algorithmic}
\usepackage{float}
\usepackage{graphicx}
\usepackage{textcomp}
\usepackage{xcolor}
\usepackage{siunitx}
\usepackage[hidelinks]{hyperref}
\usepackage{multirow} % Package for the table
\usepackage{tablefootnote} % Package for the table
\usepackage{graphicx} % For the resizebo in equation
\usepackage[portuguese]{babel}
\usepackage[firstpageonly=true]{draftwatermark}

\newcommand{\iris}{\textsc{3DR\textsuperscript{\tiny\textregistered}} IRIS}
\newcommand{\nvidia}{\textsc{NVIDIA\textsuperscript{\tiny\textregistered}}}
\newcommand{\nvidiaisaacsim}{\textsc{NVIDIA\textsuperscript{\tiny\textregistered}} Isaac Sim}
\newcommand{\nvidiaisaacgim}{\textsc{NVIDIA\textsuperscript{\tiny\textregistered}} Isaac Gym}

\newcommand{\bs}[1]{\ensuremath{\boldsymbol{#1}}} % math symbol

\newcommand{\boldcal}[1]{{\ensuremath{\boldsymbol{\mathcal{#1}}}}}% math bold calygraphy

\def\BibTeX{{\rm B\kern-.05em{\sc i\kern-.025em b}\kern-.08em
    T\kern-.1667em\lower.7ex\hbox{E}\kern-.125emX}}

\newcommand*{\defeq}{\mathrel{\vcenter{\baselineskip0.5ex \lineskiplimit0pt
                     \hbox{\scriptsize.}\hbox{\scriptsize.}}}%
                     =}  

\definecolor{matlab_blue}{RGB}{0, 113, 189}
\definecolor{matlab_red}{RGB}{217, 85, 28}
\definecolor{matlab_purple}{RGB}{126, 47, 142}
\definecolor{matlab_yellow}{RGB}{237, 177, 32}

\begin{document}

\SetWatermarkText{This paper has been accepted for publication in the\\ 2024 International Conference on Unmanned Aircraft Systems (ICUAS), \textcopyright IEEE}
\SetWatermarkColor[gray]{0.3}
\SetWatermarkFontSize{0.5cm}
\SetWatermarkAngle{0}
%\SetWatermarkHorCenter{4cm}
\SetWatermarkVerCenter{1.0cm}

\title{\LARGE \bf Pegasus Simulator: An Isaac Sim Framework for Multiple Aerial Vehicles Simulation}

\author{Marcelo Jacinto\textsuperscript{1},
	Jo{\~a}o Pinto\textsuperscript{1},
	Jay Patrikar\textsuperscript{2}, 
    John Keller\textsuperscript{2},  
    Rita Cunha\textsuperscript{1},
    Sebastian Scherer\textsuperscript{2}
    and
    Ant{\'o}nio Pascoal\textsuperscript{1}%,
\thanks{The work of Marcelo Jacinto and Jo{\~a}o Pinto was supported by the PhD Grants
2022.09587.BD and 2022.12145.BD from the Funda\c{c}{\~a}o para a Ci{\^e}ncia e a Tecnologia
(FCT), Portugal. This work was also supported by FCT, Portugal through LARSyS [DOI: 10.54499/LA/P/0083/2020, 10.54499/UIDP/50009/2020, and 10.54499/UIDB/50009/2020] and the project CAPTURE [DOI: 10.54499/PTDC/EEI-AUT/1732/2020].}
\thanks{\textsuperscript{1}M. Jacinto, J. Pinto, R. Cunha and A. Pascoal are with the Laboratory of Robotics and Engineering Systems, ISR/IST, University of Lisbon, Portugal. E-mail: {\tt \small \{mjacinto, jpinto, rita, antonio\}@isr.tecnico.ulisboa.pt}}
\thanks{\textsuperscript{2}J. Patrikar, J. Keller and S. Scherer are with the Robotics Institute, Carnegie Mellon University, USA. E-mail: {\tt \small \{jpatrika,jkeller2,basti\}@andrew.cmu.edu}}
}

\maketitle

\begin{abstract}
Developing and testing novel control and motion planning algorithms for aerial vehicles can be a challenging task, with the robotics community relying more than ever on 3D simulation technologies to evaluate the performance of new algorithms in a variety of conditions and environments. In this work, we introduce the Pegasus Simulator, a modular framework implemented as an \nvidiaisaacsim~extension that enables real-time simulation of multiple multirotor vehicles in photo-realistic environments, while providing out-of-the-box integration with the widely adopted PX4-Autopilot and ROS2 through its modular implementation and intuitive graphical user interface. To demonstrate some of its capabilities, a nonlinear controller was implemented and simulation results for two drones performing aggressive flight maneuvers are presented. Code and documentation for this framework are also provided as supplementary material.
\end{abstract}

%\begin{IEEEkeywords}
%Multi-rotor, Drones, Simulation, NVIDIA Isaac Sim, Software-In-The-Loop, PX4-Autopilot
%\end{IEEEkeywords}
% Supplementary Material Section
\section*{Supplementary Material}
\noindent\textbf{Code:} \href{https://github.com/PegasusSimulator/PegasusSimulator}{https://github.com/PegasusSimulator/PegasusSimulator}
\noindent\textbf{Site:}\hspace{0.1cm}\href{https://pegasussimulator.github.io/PegasusSimulator/}{https://pegasussimulator.github.io/PegasusSimulator}
\noindent\textbf{Video:} \href{https://youtu.be/caFPdl1rOT4}{https://youtu.be/caFPdl1rOT4}
\section{Introduction}
In recent years, there has been an increasing demand for Unmanned Aerial Vehicles (UAV)s, with a special emphasis on small multirotor systems. These rotorcraft offer an high-quality and affordable way of having a top-down view of the environment and, for this reason, it is evident why they have become the tool of choice for aerial cinematography, surveillance, and maintenance missions.

The quantum leap in the field of Machine Learning (ML) that has taken place over the past years has led to a dramatic change in the robotics landscape and in how autonomous systems are designed. In this day and age, devising intertwined control and perception systems is commonplace. The control community has long been putting efforts into combining advanced control techniques, such as Nonlinear Control, Model Predictive Control (MPC), and state-of-the-art ML methods, such as Reinforcement Learning (RL). Nevertheless, acquiring data to train and validate these algorithms can be prohibitively expensive, time-consuming, impractical, and unsafe, since using a physical vehicle to collect flight data is often necessary.

In order to provide a good environment to develop new control and motion planning algorithms, it is of fundamental importance to guarantee that the simulation framework adopted exhibits the following properties:
\begin{itemize}
    \item provide vehicle and sensor models that are physically accurate and generate data at high rates;
    \item guarantee that the simulated environment and generated multi-modal data are photo-realistic;
    \item allow the simulation of multiple vehicles in parallel;
    \item provide a simple application interface for fast prototyping;
    \item provide integration with the firmware that runs on real flight controller hardware, i.e., a combination of software-in-the-loop (SITL) and hardware-in-the-loop (HITL) capabilities.
\end{itemize}

\begin{figure}[H]
	\centering
	\includegraphics[width=0.47\textwidth]{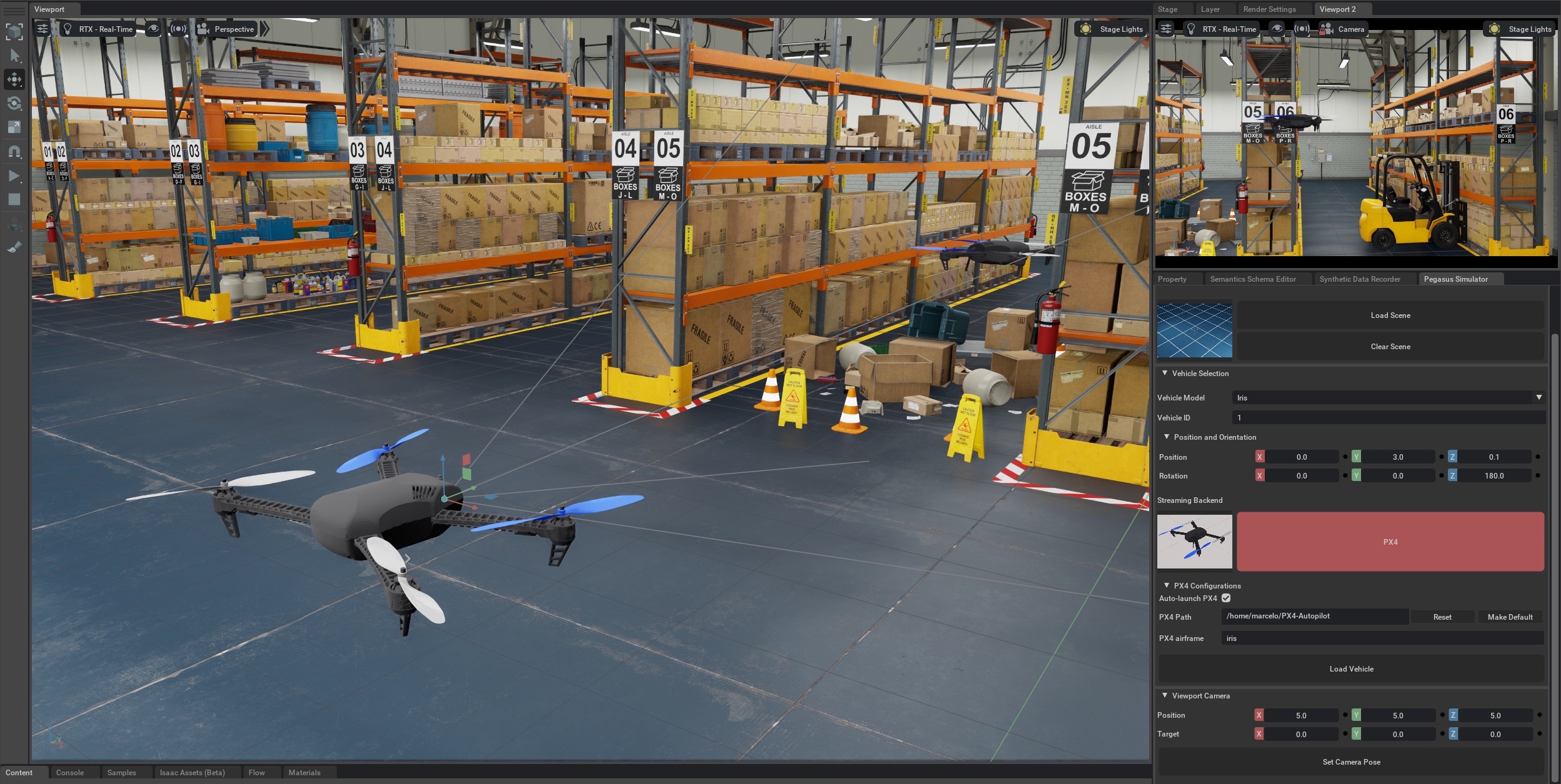}
    \vspace{-0.1cm}
	\caption{Illustration of the Pegasus Framework built with \nvidiaisaacsim, simulating two \iris~vehicles being operated in real time using the PX4-Autopilot, inside the warehouse asset world provided by NVIDIA.}
	\label{fig:simulator_complete}
\end{figure}

In this paper, we present a simulation solution that enjoys these properties -- the Pegasus Simulator, an open-source and modular framework implemented using the application \nvidiaisaacsim \cite{isaac_sim} -- with the end goal of providing a simple yet powerful way of simulating multirotor vehicles in photo-realistic environments, as shown in Fig.~\ref{fig:simulator_complete}.
\vspace{-0.1cm}
\vspace{-0.3cm}

\subsection{Related Work}

\begin{table*}[!htb]
%-----------------------------------------
% Comparison tables of existing simulators
%-----------------------------------------
\caption{Comparison of the features provided by multiple open-source simulators and multirotor simulation frameworks}
\centering
\begin{tabular}{c|c|c|c|c|c|c}
    \hline
    \multirow{2}{*}{\textbf{Simulation Framework}} & 
    \textbf{Base Simulator/} & 
    \multirow{2}{*}{\textbf{Photo-realistic}} & 
    \textbf{Onboard sensors} & 
    \multirow{2}{*}{\textbf{Vision Sensors}} & 
    \textbf{MAVLink} &
    \textbf{API} \\
    & \textbf{Rendering Engine} & & (IMU, GPS,...) & & \textbf{Interface} & \textbf{Complexity} \\
    \hline\hline
    jMAVSim          & Java    & {\color{red}x} & {\color{green}\checkmark} & {\color{red}x} & {\color{green}\checkmark} & *\\
    RotorS           & Gazebo  & {\color{red}x} & {\color{green}\checkmark} & {\color{green}\checkmark} & {\color{green}\checkmark} & **\\
    PX4-SITL         & Gazebo  & {\color{red}x} & {\color{green}\checkmark} & {\color{green}\checkmark} & {\color{green}\checkmark} & **\\
    AirSim           & Unreal Engine (or Unity) & {\color{green}\checkmark} & {\color{green}\checkmark} & {\color{green}\checkmark} & \textbf{ {\color{green}\checkmark}} & ***\\
    Flightmare       & Unity & {\color{green}\checkmark} & {\color{green}\checkmark} &  {\color{green}\checkmark} &  {\color{red}x} & ***\\
    MuJoCo           & OpenGL (or Unity) & {\color{red}x} & {\color{red}x} & {\color{red}x} & {\color{red}x} & **\\
    \hline
    \textbf{Pegasus} & NVIDIA Isaac Sim & {\color{green}\checkmark} & {\color{green}\checkmark} & {\color{green}\checkmark} & {\color{green}\checkmark} & ** \\
    \hline
    \multicolumn{7}{l}{{\color{green}\checkmark} Indicates the presence and {\color{red}x} the absence of features.} \\
    \multicolumn{7}{l}{* Indicates the degree complexity of the different frameworks, on a scale from one to three stars.}
    \end{tabular}
    \label{table:simulators_comparison}
\end{table*}

Throughout the years, several simulators and simulation frameworks have been developed, each presenting its set of strengths and weaknesses. In this section, we provide a brief overview of some of the most iconic simulation platforms that inspired this work, along with Table \ref{table:simulators_comparison}, which summarizes the main differences among them.

One of the most popular simulators used by the robotics community is Gazebo \cite{gazebo}, which provides out-of-the-box support for the widely adopted ROS framework \cite{ros} and has a rigid-body physics engine which can run decoupled from its OpenGL based rendering engine. Designed with flexibility in mind, the development of extensions that enable a combination of SITL and HITL simulations of multirotor vehicles in a virtual 3D environment, such as RotorS \cite{rotorS} and PX4-SITL (Gazebo-based) \cite{px4}, soon followed. However, Gazebo can be quite cumbersome, especially when it comes to simulating complex visual environments, and cannot compete with the visual fidelity that can be achieved with modern game engines such as Unreal Engine \cite{unreal_engine} and Unity \cite{unity}. 

This limitation has fueled the development of further simulators. One example of those is AirSim \cite{airsim} -- an extension developed by Microsoft for quadrotor simulation with a custom physics engine that interfaces with Unreal Engine for rendering and collision detection. This simulator, similarly to the PX4-SITL extension, provides basic integration with the widely adopted PX4-Autopilot and a wrapper for the OpenAI Gym~\cite{openaigym} framework targeting RL applications. There is also Flightmare \cite{flightmare}, developed at UZH, which comes with a custom physics engine and uses the Unity game engine for rendering and multi-modal sensor data generation. Similarly to AirSim, it provides integration with OpenAI Gym \cite{openaigym}, but does not provide any out-of-the-box PX4-Autopilot integration. 

Even though both simulation extensions produce visually pleasing results, they lack the simplicity that Gazebo offers, since the provided interfaces can be quite challenging to work with and extend. These solutions were built on top of game engines, not with the robotics community in mind from the ground up and, therefore, extending them is usually a difficult and slow process, as one is required to be well acquainted with not only the application interface (API) of the simulation extension, but also the complex code library of the game engine itself.

% Extending them usually requires learning not only the API of the simulation extension, but also the complex code library of the game engine itself, making it a painful and slow process due to its steep learning curve.

On the other side of the spectrum, there is jMAVSim \cite{jmavsim} -- a Java based simulator provided by the PX4 community with the sole purpose of testing the basic functionalities of the PX4-Autopilot, that runs on a variety of flight controllers such as the Pixhawk \cite{pixhawk}. Nonetheless, it does not provide any photo-realistic graphics capabilities. There is also MuJoCo \cite{mujoco}, a very efficient and general purpose rigid-body physics engine coupled with a basic OpenGL renderer, which has been popularized by its extensive use in control and RL applications, but, in contrast with previously mentioned simulators, it lacks integration with any type of existing flight control software, as it has never become widely adopted by the aerial robotics community.

Recently, \nvidia~has launched its Omniverse suite of simulation tools, which includes Isaac Sim \cite{isaac_sim} -- a modern simulator tailored for robotics applications and multi-modal data generation. This new platform provides a high-quality RTX rendering engine which can run at independent rates from its PhysX based physics engine (which is the backbone of most 3D video game titles). It also follows the Universal Scene Descriptor (USD) standard developed by Pixar and is widely supported across 3D modeling tools such as Blender, Maya and 3DS Max to represent large and complex 3D environment. In addition, it allows for the use of robot assets in URDF format adopted by the ROS community. 

The simulator also provides basic ROS1 and 2 integration, wrappers for \nvidiaisaacgim, extensions such as Isaac Orbit~\cite{orbit}, domain randomization tools used for RL applications as well as extensions to simulate humans events~\cite{humans_isaac_sim}, and multi-modal sensors such as cameras (RGB and depth) and LiDARs. Although this simulator is shipped with a very complete feature set, it lacks proper support for realistic aerial vehicle simulation. Most of the available tools are targeted at articulated robotic arms and wheeled robots development. In addition, it lacks basic sensors such as barometers and GPS and does not provide integration with popular flight controllers such as PX4 or the MAVLink communication protocol.

Taking cues from the best features of existing simulation frameworks, the Pegasus Simulator was designed as an \nvidiaisaacsim~\cite{isaac_sim} Python extension, with the end goal of providing a Gazebo-like developer experience with PX4 and MAVLink support and enabling photo-realistic simulations similar to game-engine based frameworks. The first release of the Pegasus Simulator framework provides:
\begin{itemize}
    \item a model of a \iris~quadrotor equipped with a First Person View (FPV) camera; 
    \item communications over MAVLink, with direct PX4 integration and ROS2 communications backends;
    \item a simple Graphical User Interface (GUI) embedded into the simulator to streamline the selection of 3D environment assets, vehicle models, and control backends;
    \item a Python API to seamlessly extend the framework with custom vehicle models, sensors, 3D environments, and other communication protocols.
\end{itemize}

%To solve this fragmentation in the simulation technologies available to the aerial robotics community, the Pegasus Simulator framework was designed as an \nvidiaisaacsim~\cite{isaac_sim} Python extension, with the end goal of providing a Gazebo-like developer experience with PX4 and MAVLink support while simultaneously allowing for photo-realistic simulations similar to game-engine based frameworks. The first release of the Pegasus Simulator framework provides:
\section{Software Architecture}\label{section:software_architecture}
\begin{figure}[h]
	\centering
	\includegraphics[width=0.49\textwidth]{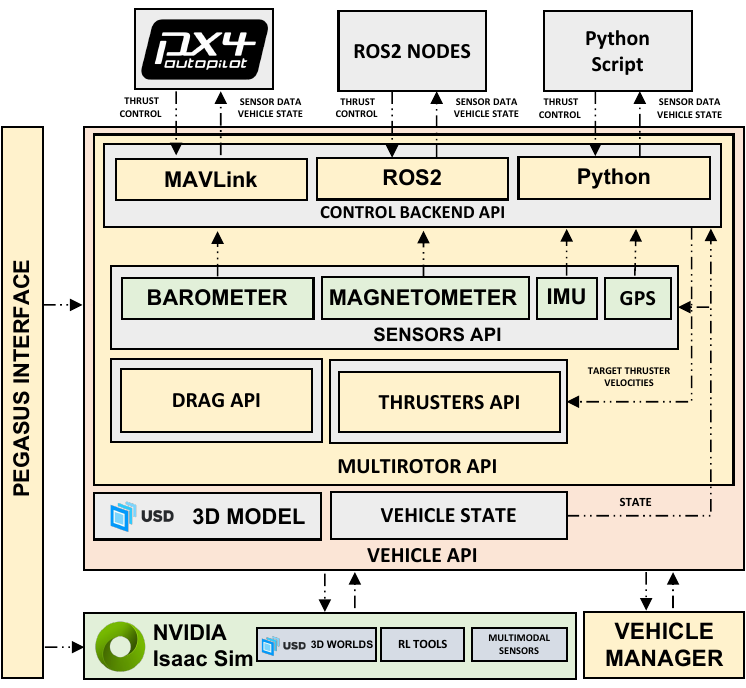}
	\caption{Pegasus Simulator architecture and its abstraction interfaces. It provides an high-level interface to \nvidiaisaacsim, a modular vehicle API that supports the addition of new sensors, thrust curves, drag models and control backends. It also provides a Vehicle Manager to keep track of all vehicles being simulated.}
	\label{fig:system_architecture}
\end{figure}

The proposed framework provides several abstraction layers that allow the creation of custom multirotor vehicles that interface directly with \nvidiaisaacsim, according to Fig. \ref{fig:system_architecture}. In this setup, a vehicle is composed by a 3D model defined in USD format that is placed in a simulation world. Following an inheritance pattern, the multirotor is a vehicle object composed of multiple sensors, a thruster model, a drag model and one or more control backends that enable the user to control the vehicle, access the state of the system and synthetic sensor data.

\subsection{Sensors, Thrust, and Drag Models}
This first iteration of the framework extends the Isaac Sim sensor suite with four additional sensors: i) barometer, ii) magnetometer, iii) IMU, and iv) GPS. The development of additional custom sensors -- via the sensors API -- is straightforward, as it provides callback functions that have direct access to the state of the vehicle and automatically handle the update rate of the sensors. It also provides abstractions for defining custom thrust and drag models. The current implementation provides a quadratic thrust curve and a linear drag model as detailed in Section~\ref{Vehicle and sensor modeling}.

\subsection{Control Backend}
In order to support a wide variety of guidance, navigation and control applications, we provide a control backend API with callback functions to access the state of the system and data generated by each onboard sensor. In turn, each control backend must implement a method to set target angular velocities to each rotor. In this layer, it is also possible to implement custom communication layers, such as MAVLink or ROS2 (already provided out-of-the-box). The implemented MAVLink backend also has the capability of starting and stopping a PX4-Autopilot simulation automatically for every vehicle, when provided with its local installation directory. This feature can prove especially convenient, as it allows the use of PX4 without having to deal with extra terminal windows. 

\subsection{Pegasus Interface}
Following this modular architecture, it is possible to not only use the provided quadrotor vehicles, but also to create new models with custom sensor configurations. The Pegasus interface provides additional tools, such as the Vehicle Manager to keep track of every vehicle that is added to the simulation environment -- either via Python scripting, or the extension GUI -- and provide access to each of them.

\subsection{Graphical User Interface}
\begin{figure}[h]
	\centering
	\includegraphics[width=0.43\textwidth]{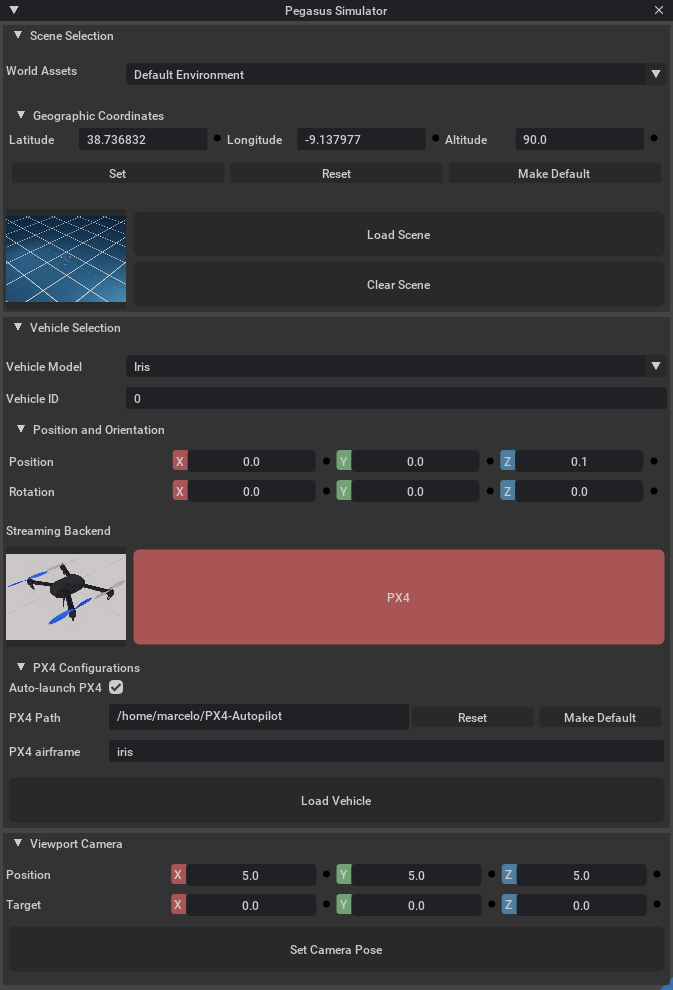}
    \vspace{-0.1cm}
	\caption{GUI provided by Pegasus Simulator extension. This visual interface allows one to select with ease a 3D simulation world from a set of available assets. It is also possible to place multiple vehicles in the simulation world and automatically start the PX4-Autopilot in the background.}
	\label{fig:graphical_ui}
\end{figure}

To simplify the simulation procedure, we provide an intuitive GUI to interact with this framework, as shown in Fig. \ref{fig:graphical_ui}. This interface allows the user to choose between a set of provided world environments and the setup of pre-configured aerial vehicles, such as the \iris~without having to explicitly work with the Python API.
\section{Vehicle and Sensor Modeling}\label{Vehicle and sensor modeling}

\subsection{Notation and Reference Frames}
\begin{figure}[h]
	\centering
	\includegraphics[width=0.39\textwidth]{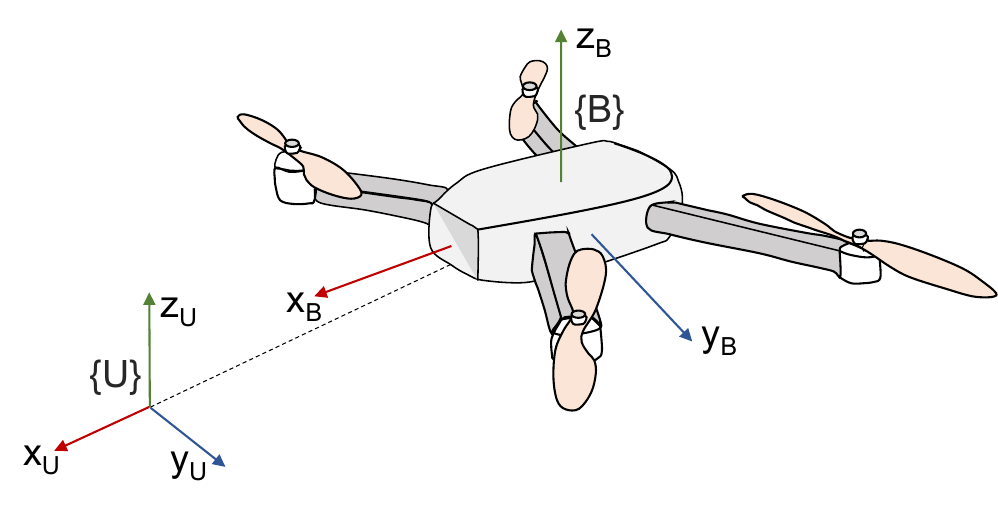}
    \vspace{-0.3cm}
	\caption{A schematic representing the vehicle body frame $\{B\}$, defined according to a FLU convention relative to an ENU inertial frame $\{U\}$.}
	\label{fig:reference_frames}
\end{figure}

Similarly to Gazebo, the Isaac Sim simulator adopts a right-handed rule convention with the Z-axis of the inertial frame facing upwards, where we arbitrate that the Y-axis aligns with true North, following an East-North-Up convention (ENU). A front-left-up (FLU) convention is also adopted for the body frame of the vehicles, according to Fig. \ref{fig:reference_frames}. 

This standard is different from the north-east-down (NED) convention for the inertial frame and front-right-down (FRD) for the vehicle body frame, adopted by the PX4-Autopilot. To maximize the support for this firmware, all the computations are performed following the ENU - FLU convention, while the data generated for all the sensors presented in Section \ref{section:sensor_modeling} is rotated in order to conform with the PX4 standard.

\subsection{System Dynamics}
We consider a dynamical model of a multirotor system affected by linear drag. Let ${\bs{p}} \defeq [ \begin{matrix} x & y & z \end{matrix} ]^\top \in \mathbb{R}^3$ denote the position of the vehicle's body frame $\{B\}$ in an inertial frame $\{U\}$ and ${\bs{v}}\defeq [ \begin{matrix} \dot{x} & \dot{y} & \dot{z} \end{matrix} ]^\top$ the velocity of $\{B\}$ with respect to $\{U\}$ expressed in $\{U\}$. The vehicle can be described by the double integrator model,
\begin{equation}
	\dot{\bs{p}} = {\bs{v}},
\end{equation}
\begin{equation}
	\dot{\bs{v}} = \underbrace{-g{\bs{e}}_{3} + {\bs{q}} \odot \frac{T}{m} {\bs{e}}_{3}}_{\eta_{1}} - \bs{q} \odot  (D \bs{q}^{-1} \odot {\bs{v}}),
\end{equation}
where $g\approx 9.81\text{ms}^{-2}$ is Earth's gravity, ${\bs{e}}_3 \defeq [ \begin{matrix} 0 & 0 & 1 \end{matrix} ]^\top$ is a unit vector, $T/m$ is the total mass-normalized thrust, and $D = \mathrm{diag}(d_x, d_y, d_z)$ is a  constant diagonal matrix with the linear drag coefficients. The operator $\odot$ is used to denote the rotation of a vector induced by a quaternion, and ${\bs{q}} \defeq [ \begin{matrix} q_x & q_y & q_z & q_w \end{matrix} ]^\top \in \mathbb{S}^3$ is the orientation of $\{B\}$ with respect to $\{U\}$. The term $\eta_{1}$ results from a combination of forces that are applied directly by the simulator physics engine on the rigid body, where the total thrust $T=\sum_{i=1}^{N} F_i$ is the sum of the forces $F_i$ applied along the Z-axis of each of the $N$ rotors. The angular velocity dynamics are expressed by
%The force resulting from the term $\eta_{2}$ is computed by the framework and manually applied to the vehicle.
\begin{align}
	\dot{\bs{\omega}} &= J^{-1} ({\bs{\tau}}- \bs{\omega} \times J \bs{\omega}), \\
    \dot{\bs{q}} &= \frac{1}{2} S_k({\bs{\omega}}) \cdot {\bs{q}},
\end{align}
where ${\bs{J}} = \mathrm{diag}(j_x, j_y, j_z)$ is the inertia tensor of the vehicle expressed in $\{B\}$, $\bs{\omega} \defeq [ \begin{matrix} p & q & r \end{matrix} ]^\top$ denotes the angular velocity of $\{B\}$ with respect to $\{U\}$ expressed in $\{B\}$, $\bs{\tau} \defeq [ \begin{matrix} \tau_x & \tau_y & \tau_z \end{matrix} ]^\top \in \mathbb{R}^{3}$ is the total torque resulting from the forces applied on each rotor and $S_k(\bs{\omega})$ is a $4 \times 4$ skew-symmetric matrix obtained from $\bs{\omega}$. The complete state of the system is given by the following quantities $\left\{ {\bs{p}},~{\bs{v}},~{\bs{q}},~\bs{\omega} \right\}$.

\subsection{Rotor Modeling}
The torque input is not applied directly on the rigid body by the physics engine, and results from
\begin{equation}
	\bs{\tau} = A {\bs{F}},
\end{equation}
where $A$ is an allocation matrix, which depends on the arm length and the position of each rotor on the vehicle, and ${\bs{F}} = [ \begin{matrix} F_1 & \cdots & F_N \end{matrix}]^\top$ is the vector of forces applied on each rotor. In particular, the z-component of the torque vector is computed according to
\begin{equation}
	\tau_z = k \sum_{i=1}^{N} (-1)^{i+1} F_i,
\end{equation}
where $k$ is the reaction torque coefficient.
In this framework, we provide a quadratic thrust model for the rotors, without time delays, given by
\begin{equation}
	F_i = c \hspace{.5ex}\omega_i^2, ~i=1,\hdots,N,
\end{equation}
where $\omega_{i}$ is the target angular velocity for each rotor provided by the control backend interface introduced in Section~\ref{section:software_architecture}, and $c \in \mathbb{R}^+$. The modular structure of the framework also allows for the development of more complex thrust models.

\subsection{Sensor Modeling} \label{section:sensor_modeling}
The Pegasus framework extends the multimodal sensor suite provided by \nvidiaisaacsim~with sensors that are commonplace in a real aerial vehicle, such as a Barometer, Magnetometer, IMU and GPS. All the implemented sensors are set to operate at $\SI{250}{\hertz}$ by default, with the exception of the GPS which is set to generate new measurements at a default frequency of $\SI{1}{\hertz}$.

%-----------------------------------
% BAROMETER
% ----------------------------------
\subsubsection{Barometer and air pressure}
According to the model of the International Standard Atmosphere (ISA), which takes into account neither wind nor air turbulence~\cite{baromter_implementation}, the temperature measurements are given by
\begin{equation}
    T = T_{0} - 0.0065 h,
\end{equation}
where $T_0 = \SI{288.15}{\kelvin}~ (\SI{15}{\celsius})$ is the Mean Sea Level (MSL) temperature and $h = z + h_0 \hspace{.5ex}[\si{\meter}]$, the vehicle altitude in meters, with $h_0$ being the altitude of the simulated world at the origin. The simulated pressure $p$ at a given altitude is given by
\begin{equation}
    p = \frac{p_0}{(T_0/T)^{5.2561}} + w + d,
    \label{eqn:pressure}
\end{equation}
where $p_0 = 101325.0 ~\si{\pascal}$, $w \sim \mathcal{N}(0, \sigma_p)$ is zero-mean additive Gaussian noise, and $d$ is a slow bias drift term computed as a discrete process. To generate pressure altitude $PA$ measurements -- which relate the rate of change of pressure with altitude -- the quotient of the additive noise term $w$ and Earth's gravity $g$ multiplied by air density $\rho = \SI{1.293}{\kilo\gram\per\meter\cubed}$ is subtracted from the true altitude of the vehicle, yielding
\begin{equation}
    PA = h - \frac{w}{g\rho}.
\end{equation}
This model is valid up to $\SI{11}{\kilo\meter}$ in altitude.

% we take the true vehicle altitude and subtract the additive noise term $w$ -- introduced in \eqref{eqn:pressure} -- divided by the Earth's gravity and air density $g\rho$, relating the rate of change of pressure with the altitude, according to

%-----------------------------------
% MAGNETOMETER
% ----------------------------------
\subsubsection{Magnetometer}
To simulate the measurements produced by a magnetometer, we resort to the same declination $D$, inclination $I$ angles, and magnetic strength $S$ pre-computed tables provided in PX4-SITL~\cite{px4} that were obtained from the World Magnetic Model (WMM-2015). From the data in those tables, the strength components $S_X$, $S_Y$ and $S_Z$, expressed in the inertial frame $\{U\}$, according to a ENU convention, are given by
\begin{equation}
    \begin{cases} 
        H &= S \cos{(I)},\\
        S_X &= H \cos{(D)} + w_X + d_X,\\
        S_Y &= H \sin{(D)} + w_Y + d_Y,\\
        S_Z &= H \tan{(I)} + w_Z + d_Z,
        \end{cases}
\end{equation}
and are computed according to~\cite{compute_mag_components}, where $w \sim \mathcal{N}(0, \Sigma_S)$ is additive white noise with covariance matrix $\Sigma_S$ and $d_X$, $d_Y$ and $d_z$ are slowly varying random walk processes.

% Good description here explaining why we do this
% https://github.com/ethz-asl/kalibr/wiki/IMU-Noise-Model

%-----------------------------------
% IMU
% ----------------------------------
\subsubsection{Inertial Measurement Unit (IMU)}
The IMU is composed by a gyroscope and an accelerometer, which measure the angular velocity and acceleration of the vehicle in the body frame, respectively. The angular rate $\boldcal{\Tilde{\omega}}$ and linear acceleration $\bs{\Tilde{a}}$ measurements are given by
\begin{equation}
    \begin{cases}
        \bs{\Tilde{\omega}} &= \bs{\omega} + \bs{\eta}_g + \bs{b}_g \\
        \bs{\Tilde{a}} &= \dot{\bs{v}} + \bs{\eta}_a + \bs{b}_a
    \end{cases},
\end{equation}
where $\bs{\omega}$ and $\dot{\bs{v}}$ are the true angular velocity of the vehicle expressed in the body frame $\{B\}$, and the first time derivative of the linear velocity expressed in the inertial frame $\{U\}$, $\bs{\eta}_g$ and $\bs{\eta}_a$ are Gaussian white noise processes, and $\bs{b}_g$ and $\bs{b}_a$ are slowly varying random walk processes of diffusion, as described in \cite{kalibr}.

%-----------------------------------
% GPS
% ----------------------------------
\subsubsection{GPS}
The local position $\bs{\Tilde{p}}$ -- used as a base for this sensor -- is given by
\begin{equation}
    \bs{\Tilde{p}} = \bs{p} + \bs{\eta}_p + \bs{b}_p,
\end{equation}
where $\bs{p}$ is the true vehicle position in $\{U\}$, $\bs{\eta}_p$ is a Gaussian white noise process and $\bs{b}_p$ a random walk process. In order to guarantee full compatibility with the PX4 navigation system, the projection from local to global coordinate system, i.e., latitude and longitude, is performed by transforming $\bs{\Tilde{p}}$ to the geographic coordinate system and using the azimuthal equidistant projection -- in accordance with the World Geodetic System (WGS84). According to this convention, the world is projected onto a flat surface, and for every point on the globe both the direction and distance to the central point are preserved~\cite{Snyder1987, azimuthal_projection}. Consider that $\phi_0$ and $\lambda_0$ correspond to the latitude and longitude of the origin of the inertial frame $\{U\}$. To generate simulated global coordinates, we start by computing the angular distance $c$ of the vehicle to the center point, according to
\begin{equation}
    c = \sqrt{x^2 + y^2},
\end{equation}
after which, the latitude $\phi$ and longitude $\lambda$ corresponding to $\bs{\Tilde{p}}$ can be computed as 
\begin{equation}
    \phi = \arcsin{\left( \cos{(c)}\sin{(\phi_0)} + \frac{y \sin{(c)}\cos{(\phi_0)}}{c} \right)},
\end{equation}
\begin{equation}
    \resizebox{.425\textwidth}{!}{$\lambda = \lambda_0 + \arctan{\left( \frac{x\sin{(c)}}{c \cos{(\phi_0)} \cos{(c)} - y \sin{(\phi_0)}\sin{(c)}} \right)}$},
\end{equation}

\noindent
if $c\neq 0$, otherwise $\phi=\phi_0$ and $\lambda=\lambda_0$.

% -------------------------------
% Old latex to be removed
% -------------------------------
% https://mathworld.wolfram.com/AzimuthalEquidistantProjection.html
%https://stackoverflow.com/questions/7222382/get-lat-long-given-current-point-distance-and-bearing
%\todo[inline]{Check this bias again. We might need to change this. Not entirely sure on the correctness of this model. PX4 mentions "based on Maybeck et al. \cite{maybeck}"}
%\todo[inline]{Finish the IMU}
%, dictated by
%\begin{equation}
%    d_{k+1} = d_k e^{-\frac{\Delta t}{\tau_{s}}} + y,
%\end{equation}
%where
%\begin{equation}
%    y \sim \Bigg (0, \sqrt{-\frac{\sigma^2}{2} \tau (e^{-2\frac{\Delta t}{\tau}} -1)} \Bigg ).
%\end{equation}
%\begin{equation}
%    d_{k+1} = d_k + \Delta t.
%\end{equation}
%\begin{equation}
%    b^{d}_{k+1} = b^{d}_{k}e^{\frac{dt}{\tau_g}} + w
%\end{equation}
%where
\section{Example Use-Case}
\begin{figure}[h]
	\centering
	\includegraphics[width=0.45\textwidth]{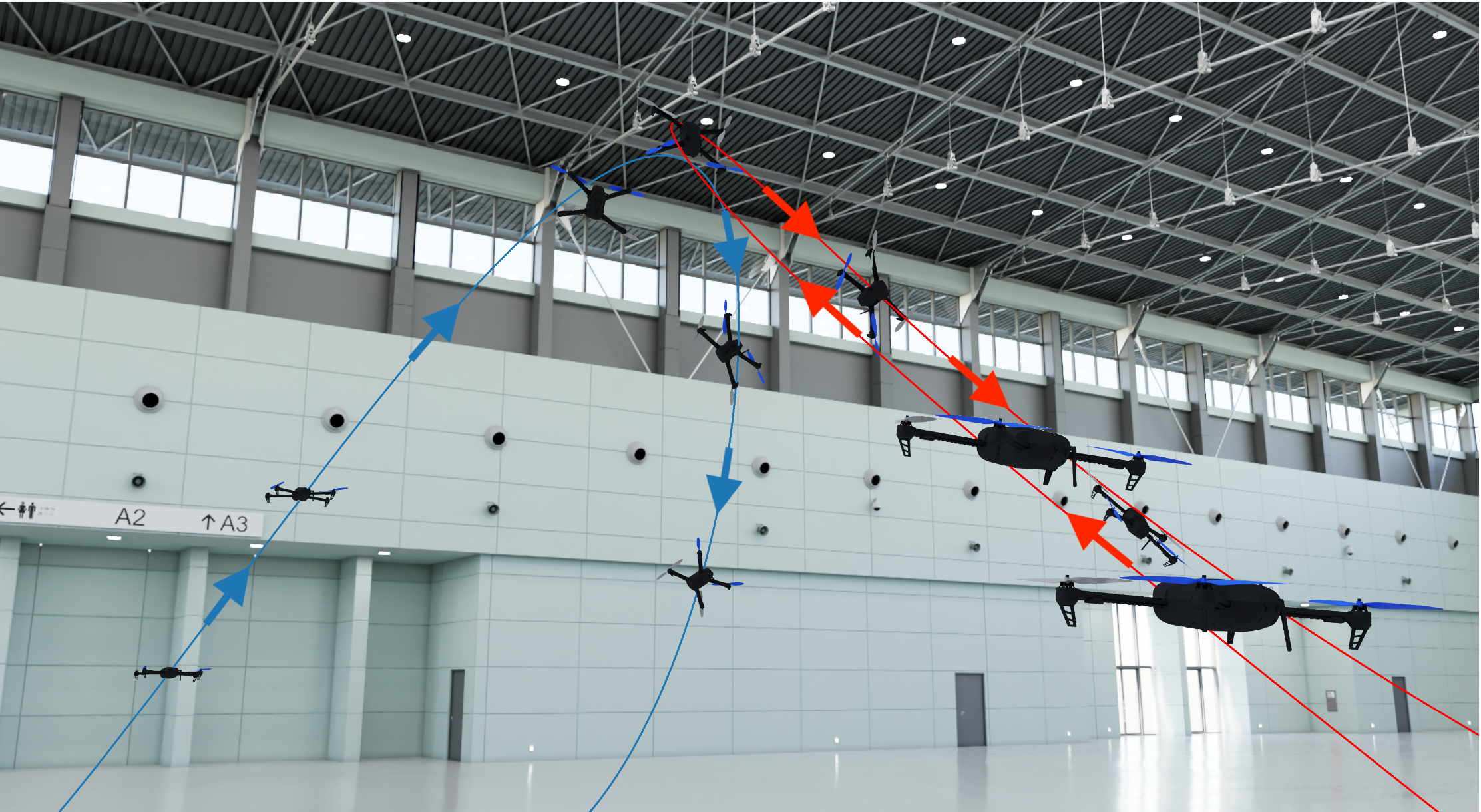}
	\caption{Composite illustration of two quadrotors performing an aggressive relay maneuver in a photo-realistic environment using a nonlinear controller implemented using the proposed 3D simulation framework.}
	\label{fig:composite_figure}
\end{figure}

To highlight the flexibility of the proposed framework, we used the control backend API detailed in Section~\ref{section:software_architecture} to implement the nonlinear controller described in Mellinger and Kumar~\cite{mellinger_kumar}. We then used the framework to replicate some of the results introduced by Pinto et al.~\cite{Pinto2021}, where two quadrotors are required to perform aggressive collision-free relay maneuvers, described by
\begin{equation}
    \begin{cases}
        x(t) &= t \\
        y(t) &= \frac{1}{s}e^{-0.5(t/s)^2} \\
        z(t) &= 1 + \frac{1}{s}e^{-0.5(t/s)^2}
    \end{cases},
\end{equation}
where $s=0.6$ is a constant that defines how aggressive the trajectory is, and $t$ is a parametric variable. This scenario, shown in Fig.~\ref{fig:composite_figure}, is particularly interesting as it would be unsafe to resort to a physical vehicle to collect flight data before prior control tuning in simulation.

\begin{figure}[h]
	\centering
	\includegraphics[width=0.45\textwidth]{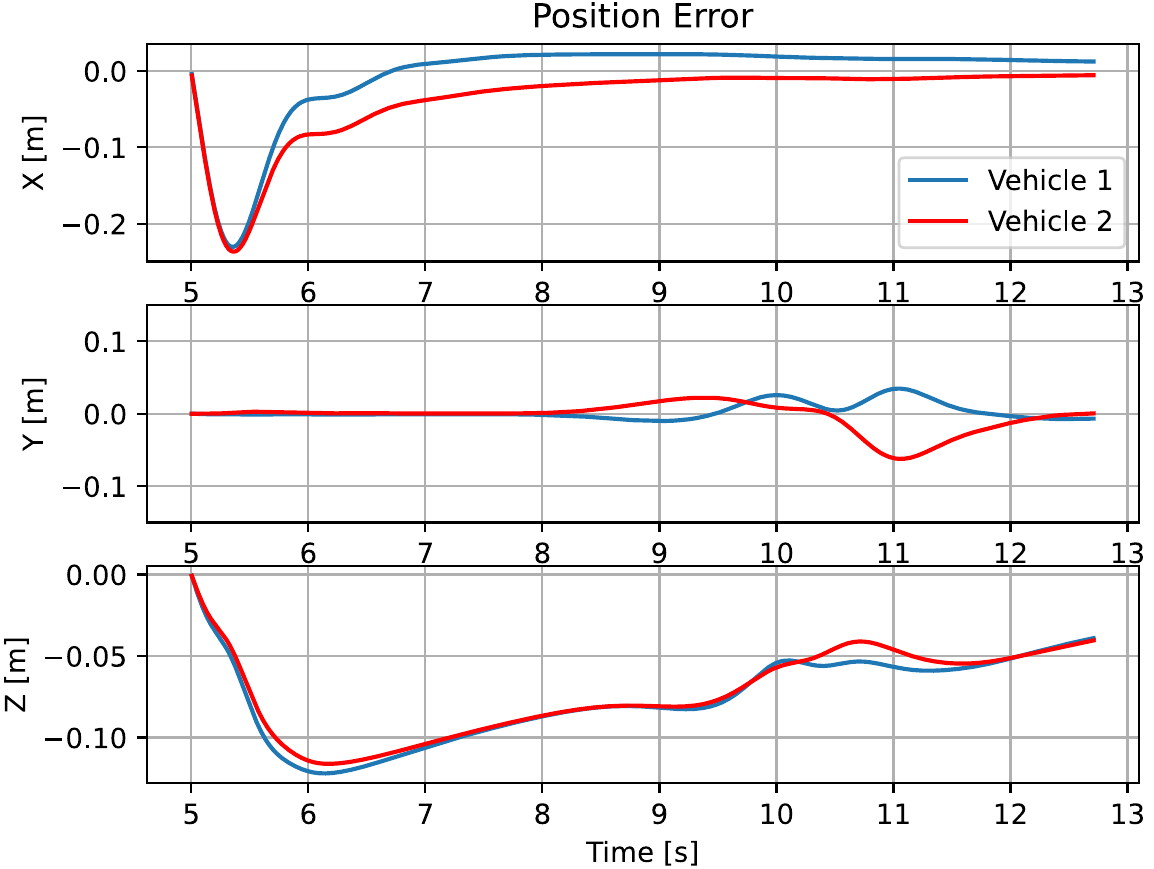}
    \vspace{-0.3cm}
	\caption{Evolution of the position tracking error in time.}
	\label{fig:tracking_error}
\end{figure}

In Fig.~\ref{fig:tracking_error} we can observe that both quadrotors are able to track their respective aggressive trajectory, with bounded position errors and similar performance characteristics to the results obtained in MATLAB~\cite{Pinto2021}. The simulation results presented were obtained on a computer equipped with an AMD Ryzen 5900X CPU and an NVIDIA RTX 3090 GPU.
\vspace{-0.4cm}
\section{Conclusion}

This paper has presented the first iteration of the Pegasus Simulator -- a framework built using \nvidiaisaacsim~for real-time photo-realistic development and testing of guidance, navigation, and control algorithms applied to multirotor vehicles. It provides an intuitive GUI within the 3D simulator for fast prototyping and testing of new control systems with the possibility of using the PX4 firmware in-the-loop and ROS2 integration out-of-the-box, which enables conducting simulations that take into account the inner workings of flight controllers found in real aerial vehicles. In addition, it provides an extensible and modular Python API that facilitates extending this framework with custom sensors, vehicles, environments, communication protocols and control layers. Its modularity streamlines the simulation of multiple vehicles at once and the use of the proposed control backend structure for custom applications. 

We believe this extension provides a middle-ground between simulators such as Gazebo -- which are feature rich and accessible to the robotics community, but not photo-realistic -- and other game-engine based simulators -- which provide impressive graphics capabilities, but at the cost of having a steep learning curve. To highlight some of the capabilities of the developed framework, an example use-case where quadrotors perform an aggressive flight maneuver has been presented. 

Future work includes the extension of this framework to other vehicle designs and an extension of the sensor suite provided, so as to improve the compatibility with PX4-Autopilot SITL, which is widely adopted by the aerial robotics community.

% This command serves to balance the column lengths on the last page of the document manually.
%\addtolength{\textheight}{-12cm}

\section*{Acknowledgment}
The authors would like to express their deep gratitude to Gil Serrano, Jo{\~a}o Lehodey, Jos{\'e} Gomes, Pedro Trindade, Bruno Guerreiro and David Cabecinhas for all the support and feedback provided during the development of Pegasus.

% Bibliography inclusion using bibtex
%\bibliographystyle{IEEEtran}
%\bibliography{IEEEabrv,bibliography}

\end{document}